# Multispectral Satellite Data Classification Using Soft Computing Approach

Purbarag Pathak Choudhury[1], Ujjal Kr Dutta[1] and Dhruba Kr. Bhattacharyya[2]
Tezpur University, Department of Computer Science and Engineering, Tezpur, India
[1]Email: {purbo002, ukdacad}@gmail.com
[2]Email: dkb@tezu.ernet.in

*Abstract*—A satellite image is a remotely sensed image data, where each pixel represents a specific location on earth. The pixel value recorded is the reflection radiation from the earth's surface at that location. Multispectral images are those that capture image data at specific frequencies across the electromagnetic spectrum as compared to Panchromatic images which are sensitive to all wavelength of visible light. Because of the high resolution and high dimensions of these images, they create difficulties for clustering techniques to efficiently detect clusters of different sizes, shapes and densities as a trade off for fast processing time. In this paper we propose a grid-density based clustering technique for identification of objects. We also introduce an approach to classify a satellite image data using a rule induction based machine learning algorithm. The object identification and classification methods have been validated using several synthetic and benchmark datasets.

*Index Terms*—multispectral satellite image, clustering, grid density, classification, rough set

## I. INTRODUCTION

A satellite image is a remotely sensed image data that records the electromagnetic radiation reflected from the surface of the earth as its pixel values. A high dimensional satellite image contains a large amount of information. Satellite images are of various types: panchromatic, multispectral, hyperspectral and ultraspectral. A satellite usually has three or more radiometers, each capturing one digital image in the small band of visible spectra. Remotely sensed satellite images mainly consist of vegetation, water bodies, open spaces, habitation and concrete structures which are distinguished by their different reflectance characteristics, leading to variety of clusters of different shapes, sizes and densities [3]. Nowadays, multispectral images are the ones mainly acquired by remote sensing (RS) radiometers.

For multispectral images, NIR (near infrared) and red wavelengths carry the greatest amount of information about the natural environment. Apart from these two, the green visible band is also included, since it can be used along with the other two to produce a traditional false colour composite (a full colour) image derived from the red, green and IR bands. Based on our study it has been observed that most algorithms fails to handle effectively handle all these factors at the same time: quality cluster detection, fast processing time and non-dependency on input parameters. Traditional automatic satellite image clustering algorithms [12] and classification algorithms are found inadequate to obtain satisfactory results in case of high resolution image sources like IKONOS, CARTOSAT I and IRS P6 LISS IV, showing a diversity of land cover forms because of high variance in the data. Furthermore, mixed pixels (mixels) are also present in satellite images. Mixels occur when more than one object on the ground are present in a single pixel of the image. SATCLUS-T [3], a grid-density based approach addresses the problem of mixed pixel handling and works in two steps: Step I which identifies the rough clusters and Step II smoothen the cluster boundaries detected in Step I.

Classification of remotely sensed data is an important aspect of remote sensing data analysis from the Land Use Land Cover classification point of view. Classification is carried out to assign corresponding levels with respect to groups with homogeneous characteristics with the aim of discerning multiple objects in an image. Popular techniques are: Minimum Distance classifier [13], Parallelpiped classifier [13][14], Maximum Likelihood classifier [13][15], Mahalanobis Distance classifier [16][17] and other classifiers such as fuzzy set classifier and expert systems. In this paper, we propose an effective multispectral satellite data classifier based on Rough Set Theory (RST). The two major attractions of our method are as follows:

i) Object identification method is robust and can identify objects of any shape i.e. concave and convex.
ii) Classification method based on RST can handle inconsistency.

These images have applications in seismology, oceanography, study of land formation, water depth and sea bed and so on. Many datasets obtained from real life problem contains arbitrary shapes. In any clustering technique it is desirable that such clusters are detected and processed.

## II. BACKGROUND

In this section we provide some basics of satellite imagery, multispectral satellite imagery, clustering and rough set theory.

### A. Satellite Imagery

Satellite imagery consists of photographs of Earth by means of artificial satellites. There are a number of satellites currently in operation which operate at various resolutions. Applications of satellite imagery range from meteorology, landscape, geology, forestry, regional planning, biodiversity conservation, education, intelligence as well as in national security and warfare. Images can be in visible colours, panchromatic and in other spectra. Interpretation and analysis of such images is conducted using specialized remote sensing applications [1].

### B. Multispectral Image

A multispectral image is one that captures image data at specific frequencies across the electromagnetic spectrum. The wavelengths are separated by filters that are sensitive to particular wavelengths. Multispectral images include light from frequencies beyond the visible light range, such as infrared. Spectral imaging allows extraction of additional information the human eye fails to capture with its receptors for red, green and blue. It was originally developed for space-based imaging.

Each of three or more radiometers in a satellite acquires one digital image (a 'scene' in remote sensing) in a band of visible spectra ranging from 0.7 to 0.4 μm called RGB (red-green-blue) region and going to infrared wavelengths of 0.7 to 0.4 μm or more, classified as NIR (near infrared), MIR (middle infrared) and FIR (thermal or far infrared) [2]. The information is stored in the satellite images in the form of spectral band data. Spectral bands are well-defined, continuous wavelength range in the spectrum of reflected or radiated electromagnetic energy. Various spectral bands are [2]:

i) Blue, 450-515..520 nm, atmospheric and deep water imaging
ii) Green, 515..520-590..600 nm, imaging of vegetation and deep water structures
iii) Red, 600...630-680...690 nm, man-made objects, soil, and vegetation
iv) Near infrared, 750-900 nm, primarily for imaging of vegetation.
v) Mid-infrared, 1550-1750 nm, vegetation, soil moisture content, and some forest fires.
vi) Mid-infrared, 2080-2350 nm, soil, moisture, geological features, silicates, clays, and fires.
vii) Thermal infrared, 10400-12500 nm, uses emitted radiation instead of reflected, for imaging of geological structures, thermal differences in water currents, fires, and for night studies.

Mapping of bands to color is done for the purpose of imaging. Thermal infrared is often omitted from consideration due to poor spatial resolution, except for some special purposes. Different combinations used are:

i) True-color uses only red, green, and blue channels, mapped to their respective colors. As a plain color photograph, it is good for analyzing man-made objects, and is easy to understand for beginner analysts.
ii) Green-red-infrared, blue channel is replaced with near infrared, used for detection of vegetation and camouflage.
iii) Blue-NIR-MIR, where the blue channel uses visible blue, green uses NIR (so vegetation stays green), and MIR is shown as red. Such images allow seeing the water depth, vegetation coverage, soil moisture content, and presence of fires, all in a single image.

Applications of spectral imaging include astronomy, solar physics, analysis of plasmas in nuclear fusion experiments and so on.

### C. Object Recognition By Clustering And Segmentation

Cluster analysis of a dataset is an unsupervised method [18] for exploring the underlying structure of a given dataset and organizing the dataset into some well separated partitions with respect to certain similarity measures. It is widely used in detection of object and pattern recognition technique applied in numerous domains such as business intelligence, image pattern recognition, Web search, biology and security. Clustering, also sometimes called data segmentation, is a form of learning by observation rather than learning by examples. Several clustering approaches are in existence to provide grouping and classification of any data from real life domain. However the four fundamental clustering approaches are briefly introduced next [11]:

i) *Partitioning method*: Given a set of N objects, a partitioning method constructs k partitions of the data such that $k \leq N$. Most partitioning methods are distance-based. After creating an initial set of partitions it uses an iterative relocation technique to improve the quality of partitions. These methods find mutually exclusive spherical clusters.

ii) *Hierarchical method*: Hierarchical clustering methods decompose the data into hierarchies of clusters. They can be classified as agglomerative or divisive. While merging or splitting the clusters, a hierarchical method can use distance neighborhood or density-based.

iii) *Density based method*: Density-based method follows a density notion to divide a set of object into multiple exclusive clusters. These methods can identify clusters of any shape and can be extended from full space to subspace clustering. The main advantage of this method is that they can handle noise and can cluster any numeric dataset naturally. The idea is that density of points within a cluster is higher than those outside of it. DBSCAN[4] is such an example. It has a few drawbacks since it is difficult to estimate appropriate values of distance (ξ) and the minimum number of points (MinPts) for different datasets and the neighborhood query leading to high computational complexity (expensive in case of large datasets).

iv) *Grid based method*: The grid-based method works by quantizing the object space into a finite number of cells and performing operations on the finite set. Scalability and handling of noise are two major attractions of this approach. These methods divide the data space into a finite number of grid cells on which clustering could be performed. It has many advantages as the number of cells is independent of the number of data objects and is insensitive to the order of input data points. STING[5] and WaveCluster[6] are such examples that uses multi-resolution approach and helps in detecting clusters of data at varying levels of accuracy.

v) *Grid density based method*: Among the various clustering approaches, the density based approach is very popular due to its robustness against noise and outliers. Grid-based approach is known for its faster processing time. SATCLUS-T[3] is a grid-density based method which initially divides the image is divided into equal sized grids based on (x,y) coordinates. The maximum occurring band value (hue value) is used to choose a grid cell as the 'seed' for cluster expansion. The grid cells are clustered by searching and merging neighbor cells that have similar pixels in terms of band values until further expansion is possible. The method then restarts with next maximum occurring band value from the unclassified grids. When no further expansion is possible, the process terminates and proceeds for expansion of next cluster.

To evaluate any clustering technique, the ability to identify any complex shaped clusters and handling of different data types of high-dimensional data are two important requirements.

*D. Rough Set Theory*

Most often high dimensional data like multispectral satellite images consist of inconsistencies. An inconsistency refers to a situation when we have a different "decision" for two or more examples that have a similar combination of attribute values. Although different rule generation approaches such as frequent association rule mining, rare association rule mining, multi objective rule mining are available to support rule-based decision support activity. However these approaches are not adequate in handling inconsistencies [8].

*E. Motivation*

In this work, we aim to handle any multispectral data of any size and inconsistencies with the objective of:
i) Identifying any object of interest.
ii) Classifying such multispectral satellite objects with a rule-based approach using RST.

While generating the rules, it is also aimed to identify the minimal rules. Since regularities hidden in data are often expressed in terms of rules, rule induction is one of the most important techniques of machine learning, especially in classification. Usually rules are expressed as follows [8]:
If $(a_1, v_1), (a_2, v_2),…,(a_k, v_k)$ are (attribute, value) pairs, where $a_1, a_2,…,a_k$ are attributes, $v_1, v_2,…,v_k$

are values and (d,w) is the set of decision-value pair, then possible rule types are:

i) $(a_1, v_1) \wedge (a_2, v_2) \rightarrow (d, w)$

ii) $(a_4, v_4) \wedge (a_7, v_7) \rightarrow (d, w)$

iii) $(a_i, v_i) \wedge (a_j, v_j) \wedge (a_m, v_m) \rightarrow (d, w)$

## III. Problem Formulation

For a given multispectral image I of n objects i.e. $O_1, O_2, \ldots, O_n$ the problem is to
i) Identify each object of any shape accurately
ii) Based on the result of clustering recognize the class of each object with best possible accuracy
iii) Classify the image based on RST approach.

## IV. Proposed Work

Figure 1 shows the architecture of the proposed work. The clustering technique of our method is a variant of SATCLUS-T. Based on our experimental study it has been observed that pixel clustering is a global analysis of colour space. Therefore, instead of finding seed cell in each pass and conducting a topological neighbor searching around seed cells, we in our technique conduct topological neighbor searching around seed cells and all the cells that match with the value of the seed pixel for that pass. This modification has led us towards significant improvement in the rough clustering phase of SATCLUS-T to eliminate the need for similarity search and fining seed cell for each pass. Instead of merging cells we execute the clustering over passes. For each pass there is a pixel that has currently the highest band value. Then over consecutive traversal through unclassified cell space, cells are assigned to a cluster if it has a certain percentage of pixels (determined by a parameter called gamma, ɣ) of hue value that matches with the highest band value. Ө is a user parameter that is used to maintain the difference in band values for two pixels. The proposed method uses the cluster results while populating the rule base by creating a file containing the band values of the pixels in each cluster. For creating the classification labels, human intervention is required so as to construct a meaningful and good classified file. This file is given as input along with the image to be classified to the classification process. The process uses the file to construct the rule table by using the RST based approach and classify the given image pixel wise thereby producing the image of classification.

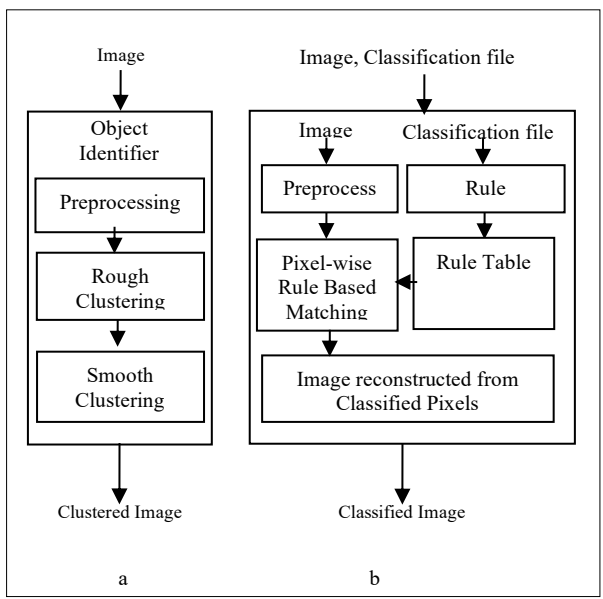

Figure 1: Architecture of the proposed system. Figure (a) shows the object identification and clustering process. Figure (b) shows the process of generating rule and classification of the image based on the rules.

## V. Related Work

In our work band values are expressed using the HSI colour model as it is the most suitable colour model for human interpretation [7]. Below we define some of the concepts in the light of [3] used in the work:

Definition 1: *Density of a grid* is defined as the total number of pixels within a particular grid cell.

Definition 2: *Seed pixel* in the image for each pass is defined as the pixel with the highest band value for the current pass.

Definition 3: Difference of HSI values (Euclidean, Manhattan or Mahalanobis distance) of a pixel with respect to the seed pixel is set to 1, if it is less than or equal to Ө. Else it is set to 0. For our work, we used Manhattan distance to find the difference value.

Definition 4: *Population count of a grid* is defined as the number of ones in a grid.

Definition 5: *Population object ratio* is the ratio of the population count and grid density.

Definition 6: *Seed grid cell* for a pass is the grid cell that has the highest population object ratio with respect to a seed pixel for the current pass.

Definition 7: *Border grid cell* is the cell which is part of a cluster $C_i$ and at least one of its neighbors is part of another cluster $C_j$.

### A. Criteria for Rough Cluster Formation

Population object ratio of a grid cell is considered as the percentage of pixels present in it that are similar to the seed pixel with respect to band value.

Therefore for a given pass, if a grid cell has $\leq Y\%$ of pixels that are similar to the seed pixel, then that grid is assigned to the cluster of the seed cell. $Y = \theta \times P_o$, where $P_o$(seed) is the population object ratio of the seed cell.

*B. Theoretical Background on Rough Set Theory*

Rough Set is an approach to vagueness i.e. imprecision or lack of knowledge in data. It focuses on the concept of boundary region of a set as compared to Fuzzy sets that uses the partial membership concept. RST is defined based on the following concepts.
i)  Universe(U): It is a given set of objects
ii) Indiscernibility relation: It is an equivalence relation
$$R \subseteq U \times U$$
that represents the lack of knowledge about elements of U.
iii) Boundary Region approach: It defines the existing of objects which cannot be classified to the set or its complement.
a. Boundary region = $\phi$ implies that set is crisp (precise).
b. Boundary region $\neq \phi$ implies that the knowledge about the set is insufficient to define it precisely.

Now suppose $X \subseteq U$; to characterize X w.r.t R, it defines the following concepts:
i)  R-Lower approximation (of X w.r.t R): It is the set of all objects certainly in X w.r.t R.
$$R_*(x) = \cup_{x \in U} \{R(x): R(x) \subseteq X\}$$

ii) R-Upper approximation (of X w.r.t R): set of all objects possibly in X w.r.t R.
$$R^*(x) = \cup_{x \in U} \{R(x): R(x) \cap X = \}$$

iii) R-Boundary region of X: is defined as
$$RN_R(X) = R^* - R_*$$

iv) Concept/Granule: It is equivalence classes of the indiscernibility relation. They are the elementary building blocks of knowledge.

LEM2 [8][9] (Learning from Examples, Module 2) is a machine learning algorithm used to generate decision rules to classify objects. Our reasons for choosing LEM2 based approach to classify the satellite data are as follows [10]:
i)  It can learn a discriminate rule set, i.e. smallest set of minimal rules describing the concept.
ii) Independent of the order of input.
iii) Can generate both
   (a) Certain rules (from lower approximation of concepts) and
   (b) Possible rules (from upper approximation of concepts).
iv) Suitable for rule generation for inconsistent data (provided the algorithm is given a lower/upper approximation or a concept itself).

LEM2 is defined based on the following concepts:
i)  U = Set of examples
ii) C = Concept to be learned, $C \subseteq U$, represented by (d,w): (decision, value) pair. Each value of d describes a unique concept C.
   a) Positive examples: The examples drawn from C.
   b) Negative examples: The examples from
$$C^c = U - C.$$

iii) Description D (of concept C): It is
   a) Expressed by a set of rules/decision trees (easier to comprehend for humans).
   b) Usable by computer programs, e.g. expert systems.

iv) Production Rule: The rules such as $C^1 \wedge C^2 \wedge C^3 \wedge ... \wedge C^i \rightarrow A$,

where $C^1, C^2, C^3, ..., C^i$ are conditions, presented as (a,v) and A is the action are referred to as production rules which is also presented as (d,w).

iv) [t]: set of all examples that for attribute 'a' have value 'v', i.e. block of attribute-value pair t = (a,v).
v)  (d,w): set of all examples that have value w for decision d.
vi) T: a set of attribute value pairs.
vii) C depends on
$$T \Leftrightarrow \neq \cap_{t \in T} [t] \subseteq C.$$

viii) Minimal Complex: T is a minimal complex of C if and only if C depends on T and T is minimal.

ix) Local Covering ($\tau$): A non-empty collection of set of attribute-value pairs.
$\tau$ is a local covering of C if and only if:
a) Each member of C is a minimal complex of C.
b) $\cup_{T \in \tau} = B$, where B is a non-empty lower/upper approximation of a concept represented by (d,w) pair or a concept itself.

c) τ is minimal.

For rule generation, we have used the LEM2 based approach based on the clustering results obtained in the previous phase.

## V. OBJECT IDENTIFICATION USING GRID DENSITY BASED APPROACH

The clustering approach adopted to identify the objects in multispectral satellite image space over HSI domain is reported phase wise next. In the first phase, it identifies the rough object patterns (i.e. clusters or segments) and the second phase attempts to smoothen the boundaries of the object identified.

### A. Object Identification using Rough Phase

i) Create the grid structure by dividing the image into $n \times n$ grids;
ii) Compute the density of each cell;
iii) Convert the RGB values of each pixel into their HSI values;
iv) Find the highest hue value among the unclustered pixels, that pixel is chosen as the seed pixel;
v) The population count of each grid cell is computed and the corresponding population-object ratio is calculated;
vi) Among the unclustered cells the one with the highest population object ratio is chosen as the seed cell. It is assigned a cluster;
vii) Traverse all the remaining unclustered cells and if their population object ratio is greater or equal to ɣ assign them to the cluster of the seed cell.
viii) Repeat steps 4 through 7 till all cells are clustered;

### B. Handling Object Boundary

For each object identified in the previous phase, in this phase we attempt to obtain a set of finer clusters based on the outputs of the previous phase. It takes arbitrarily a pixel in the border of an object and tries to judge if best possible relevant cluster based on a distance-neighborhood approach, validated by a well-established homogeneity measure called 'β-measure'. It repeats this process for all the border pixels of all the objects. However, it saves the cost of border pixel handling by avoiding unnecessary checking or validating those pixels which are distance-wise already nearer (inside) than classified distant neighbors w.r.t. a given seed pixel. We enumerate the execution of the logic for m distinct objects identified in the previous phase as follows:

i) For each object $O_i$ represented by seed pixel $P_i$ do

ii) For each border pixel $x_j^{O_i}$ do

  a) Compute dist($x_j^{O_i}$, $P_i$)

  b) If the dist is found less than its classified neighbor border pixel w.r.t. $P_i$ then assign class id of the neighbor border pixel to $x_j^{O_i}$.

  c) Else, compute β-measure for $O_i$ with $x_j^{O_i}$.

  d) If the value of β-measure is less while comparing to any other objects $O_{i+k}$ with $x_j$, then assign class id of $O_i$ to $x_j^{O_i}$.

iii) Assign $x_j$ to the cluster to which it has minimum distance with respect to the seed;
iv) Repeat steps 1 to 4 till all border pixels have been reassigned;

### C. The Mixed Pixel Problem

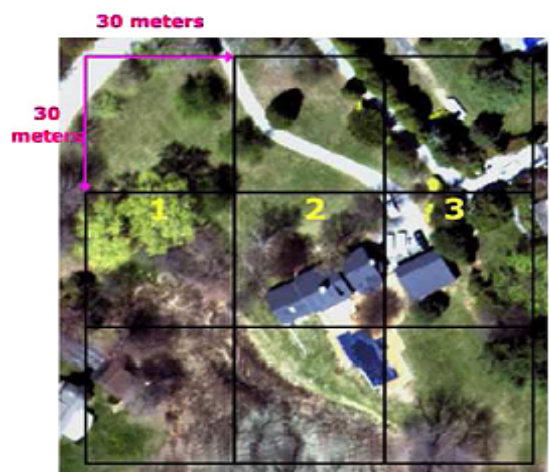

Figure 2: Image courtesy [3]

i) Pixels in the image covered by more than one classes on the ground are mixed pixels.
ii) This problem occurs because the spatial resolution of satellite sensor systems imaging the earth is coarser than sizes of objects on the ground [3].
iii) So Pixels usually cover more than one object on the ground.
iv) They greatly affect the quality of clusters produced.

e.g. In Fig. 2 above, pixel 3 is a mixed pixel since it represents many non-homogeneous objects like concrete structures, vegetation, roads and vehicles. Fig. 3 below shows how different cases of mixels can occur,

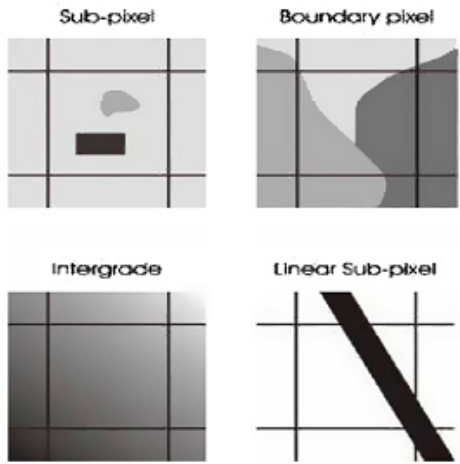

Figure 3. Different cases of mixed pixels. Image courtesy [3]

Based on our experience of handling several multispectral satellite images, it has been observed that after applying our grid-density-based clustering technique there may be a necessity of human intervention, especially during the database population to construct meaningful database objects based on cluster results.

## VI. CLASSIFICATION USING RST BASED APPROACH

In this phase we use RST approach to classify the objects identified by previous phase. The classification process includes three basic steps:

### A. Finding Local Cover ($\tau$)

Input: a concept C
Output: the local covering $\tau$

Steps:
i) Select different concept, i.e. find all [(d,w)].
ii) Find all [t], i.e. blocks of attribute-value pairs.
iii) Select the required concept. Let it be C.
iv) Set initial goal G=C.
v) Find the set T(G): set of attribute-value pairs relevant with G, i.e. pairs whose blocks have non-empty intersection with G, i.e.
$$T(G) = \{t | [t] \cap G \neq \emptyset\}$$
vi) Find the most relevant attribute-value pair in $T(G)$, i.e. the one with maximum cardinality of $[t] \cap G$. If a tie occurs, choose the one with minimum cardinality of $[t]$. If further tie occurs, choose the first attribute-value pair.
vii) Check if there exists any minimal complex T of C that is computed from relevant attribute-value pairs from $T(G)$, for the current iteration. Using the minimal complexes find the local covering $\tau$.

viii) Update G as: $G = G - \{[T] | t \in \tau\}$.

ix) Repeat steps 5 to 8 while $G \neq \emptyset$.

### B. Constructing Rule Table

Let $\tau = \{T_1, T_2, T_3, ..., T_i\}$, where

$T_1 = \{t_1, t_2, t_3\}, T_2 = \{t_4, t_5\} ... T_i = \{t_k\}$ (say) and

$t_1 = (a_1, v_1), t_2 = (a_2, v_2) ... t_k = (a_n, v_n)$,

Then, the minimal set of rules the concept (d, w) is:
$(a_1, v_1) \wedge (a_2, v_2) \rightarrow (d, w)$
$(a_4, v_4) \wedge (a_6, v_6) \rightarrow (d, w)$
$(a_i, v_i) \wedge (a_n, v_n) \rightarrow (d, w)$

These minimal cover rules are used to populate the rue table for classification.

### C. Classification

For each object identified in a given input image, the pixel values are stored in a one dimensional array. To classify the class id of each object, we match the pixel values of the corresponding array element against the rule base. Based on match result class label is assigned to each object.

## VII. RESULTS AND DISCUSSION

The synthetic images used for testing purpose were created using the Microsoft Paint program. The testing was carried out on a system with Intel Core2 Duo processor with Windows 7 operating system and 3GB RAM.

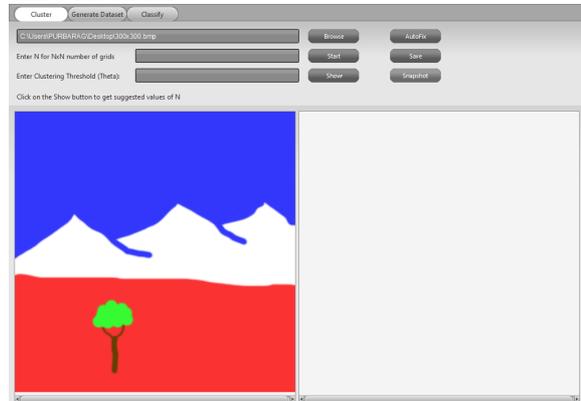

Figure 4: Interface of our system

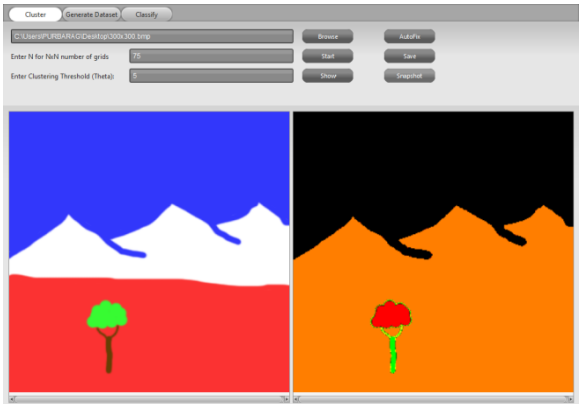

Figure 5: Result of clustering of image in Fig. 4

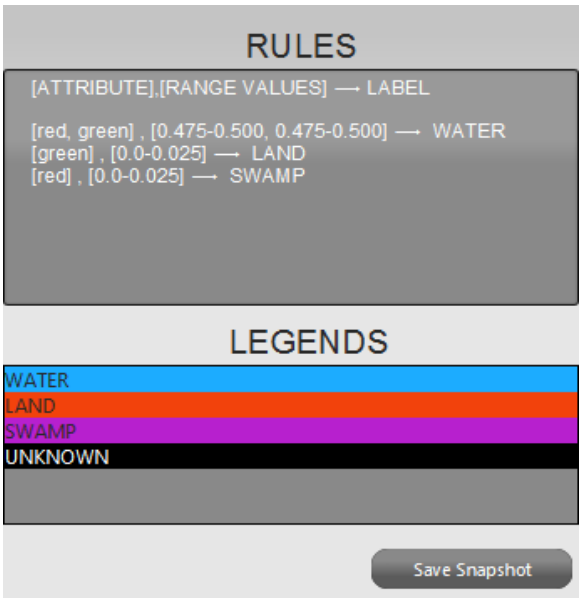

Figure 6: GUI showing the rule table.

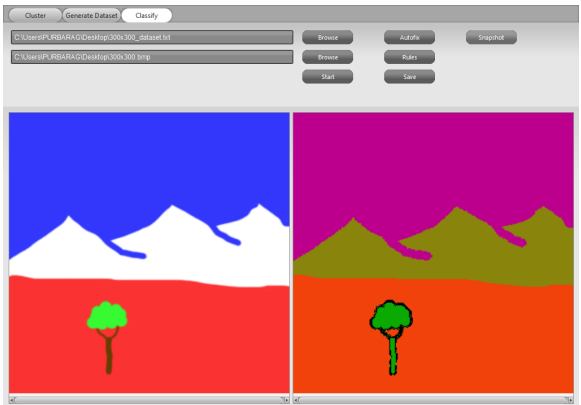

Figure 7: Result of classification of image in Fig 4

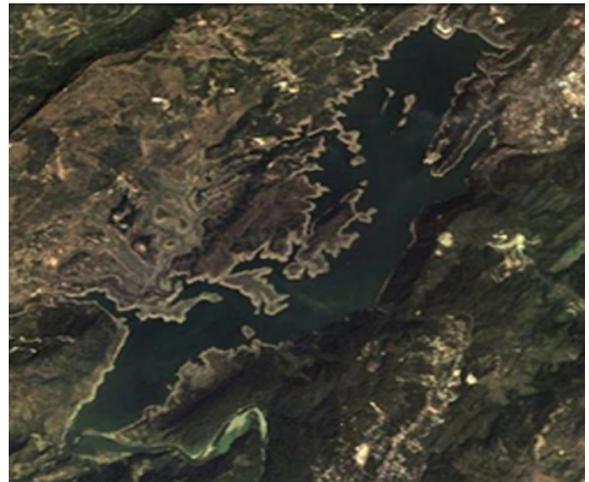

Figure 8: Multispectral satellite image of Borapani lake, Shillong, Meghalaya

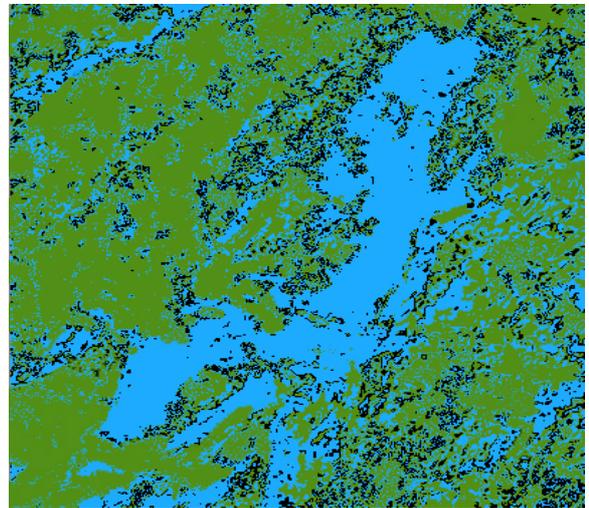

Figure 9: Result of classification of image in Fig. 8.

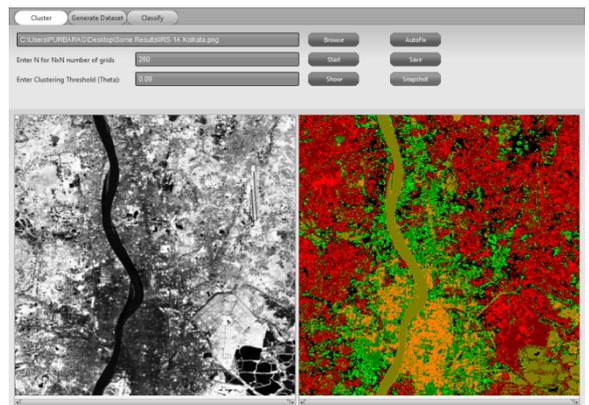

Figure 10: Panchromatic image of IRS, Kolkata along with the result of clustering

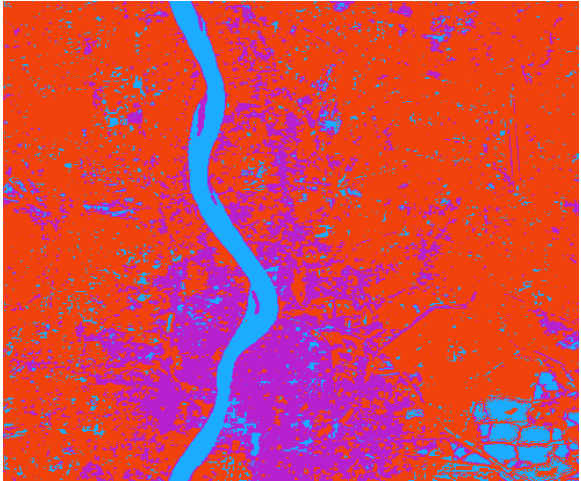

Figure 11: Result of classification of image in Fig. 10 by the generated rule in Fig. 6

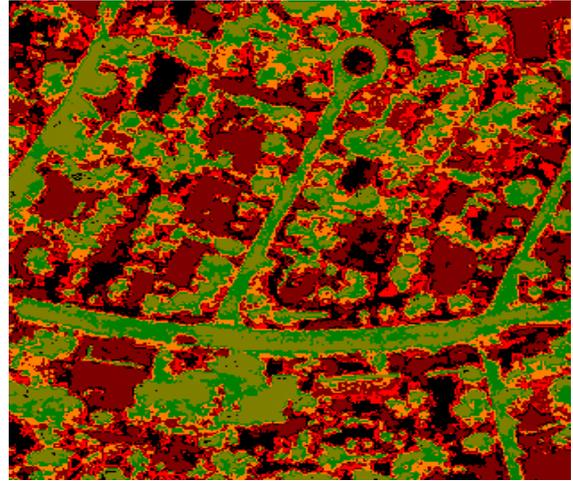

Figure 14: Image in Fig. 12 clustered by our technique

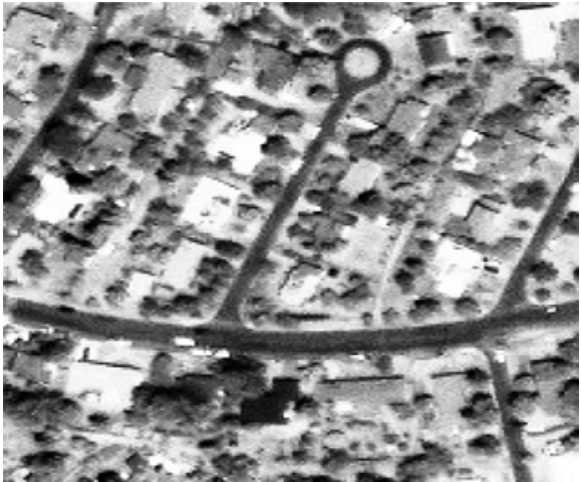

Figure 12: A panchromatic image of Mexico. Image courtesy: [13]

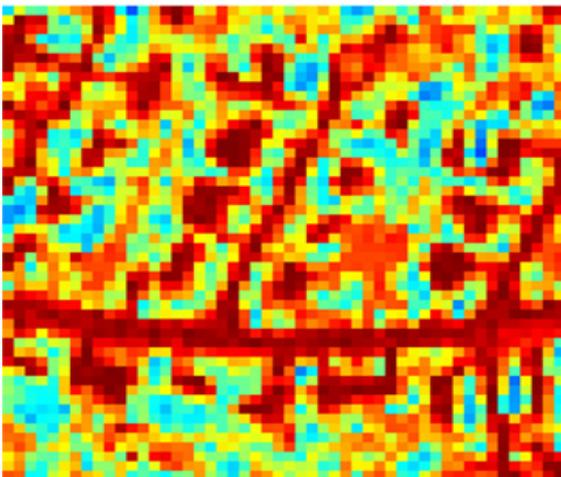

Figure 13: Image in Fig. 12 clustered by 'K-means clustering with spatial coherence'. Image courtesy: [13]

Fig. 4 shows the GUI of our interface along with one of the synthetic images used for testing. Fig. 5 and Fig. 7 show the results of clustering and classification on the synthetic image. It can be observed that the both the clustering as well as classification techniques proposed in this paper successfully identified objects of concave and convex shapes, which is further asserted by the in the other Fig. 9, Fig. 11 and Fig 14. The rules which are generated after inputting a classification file as shown in Fig. 1 are displayed on a GUI as in Fig. 6. Fig. 8 and Fig. 9 show the multispectral satellite image and its classification output respectively. The result of clustering panchromatic image is shown Fig. 10. Fig. 11 shows the result of classification of the image in Fig. 10 using the rule set generated which are shown in Fig. 6. Fig. 12 is a panchromatic image of Mexico used to show the difference in the clustering results which are in Fig. 13 and Fig. 14. It can be observed that our proposed technique gives better result for clustering.

After running the clustering algorithms over few synthetic images and multispectral satellite images the following have been observed:

i) If Ɵ is very less, so is ɣ. This inhibits the cells to be assigned to a cluster easily leading to more processing time. But due to small Ɵ there is an advantage that different shades of the same colour would be distinguished more properly. From further experimentation, it was found that smaller value for Ɵ and larger value for grid number tends to give better result. However it is not true for all cases.

ii) If Ɵ is very high, effective number of clusters formed decreases as different shades of the same colour is considered as a single color. For example, cases may arise where a pink and a red cluster are merged to form a single cluster.

iii) The value of grid size must be small enough so as to properly utilize the advantages of grid-based clustering approach. If it is taken too large, the rough clustering phase would nearly become pixel level clustering leading to more processing time.
iv) For different images, different values of Ө and grid size would lead to different quality clusters. Best values would have to be searched exhaustively.
v) Most importantly, this technique has been able to correctly detect clusters of any convex as well as concave shapes.

## VII. Complexity Analysis

The clustering phase largely depends on the input of the number of grids, N. If the N is too large the Rough Clustering phase will become equivalent to the Smooth Clustering phase and hence the algorithm take more time to produce the output. On the other hand if N is smaller, the speed of the algorithm increases at the cost of degraded output. Hence the complexity of object identification phase of clustering technique is $O(N \times K)$, where $K$ = number of seed pixels. During handling of object boundaries, the identification of the $Q$ border grid cells requires $O(Q)$ times. The assignment of $R$ pixels to $K$ clusters requires $O(K \times R)$ times, where $R$ is the total number of pixels in $Q$ border cells. Therefore, total time complexity for handling of object boundaries is $O(Q) + O(K \times R)$.

Therefore, the overall complexity of clustering technique is $O(N \times K) + O(Q) + O(K \times R)$.

In the classification phase, the complexity of the rule generating algorithm, LEM2 is $O(N \times M)$ where $N$ is the number of training data instances and M is the number of attributes. The classification is done by comparing each pixel value vector of the input image with the value vector of each rule. Therefore, the complexity of classification is $O(K \times L)$ where $K$ is the number of pixels in the image and L is the number of rules generated by LEM2. Hence the overall, complexity of the classification phase is $O(N \times K) + O(K \times L)$.

## VIII. Conclusion and Future work

This paper has presented an effective grid density-based clustering technique to identify objects of any convex as well as concave shapes on multispectral satellite images. The method has been established to perform satisfactorily on several real life multispectral satellite images. We have also introduced a classification approach based on rough set theory approach that uses the objects identified by the clustering technique. The performance of the classification method has been found satisfactory on several test datasets.